\def\year{2021}\relax
\documentclass[letterpaper]{article} 
\usepackage{aaai21}  
\usepackage{times}  
\usepackage{helvet} 
\usepackage{courier}  
\usepackage[hyphens]{url}  
\usepackage{graphicx} 
\usepackage{natbib}  
\usepackage{caption} 
\usepackage{amsmath}
\usepackage{amsmath}
\usepackage{amssymb}
\usepackage{booktabs}
\usepackage{multirow}
\usepackage{diagbox}
\usepackage{bbm}
\usepackage{xcolor}
\usepackage{comment}

\usepackage{amsthm}
\theoremstyle{definition}

\title{Factorized Deep Generative Models for Trajectory Generation with Spatiotemporal-Validity Constraints}
\author{
    Liming Zhang\textsuperscript{\rm 1}, Liang Zhao\textsuperscript{\rm 1}, Dieter Pfoser\textsuperscript{\rm 1}\\
}
\affiliations{
    \textsuperscript{\rm 1}
    George Mason University\\
    4400 University Dr,\\
    Fairfax, Virginia 22030\\
    \{lzhang22, lzhao9, dpfoser\}@gmu.edu
}

\begin{document}

\maketitle

\thispagestyle{plain}
\pagestyle{plain}

\begin{abstract}
Trajectory data generation is an important domain that characterizes the generative process of mobility data. Traditional methods heavily rely on predefined heuristics and distributions and are weak in learning unknown mechanisms. Inspired by the success of deep generative neural networks for images and texts, a fast-developing research topic is deep generative models for trajectory data which can learn expressively explanatory models for sophisticated latent patterns. This is a nascent yet promising domain for many applications.
We first propose novel deep generative models factorizing time-variant and time-invariant latent variables that characterize global and local semantics, respectively. We then develop new inference strategies based on variational inference and constrained optimization to encapsulate the spatiotemporal validity. New deep neural network architectures have been developed to implement the inference and generation models with newly-generalized latent variable priors. The proposed methods achieved significant improvements in quantitative and qualitative evaluations in extensive experiments.
\end{abstract}







\section{Introduction}
Recent advances in Global Positional System (GPS), traffic surveillance cameras, unmanned aerial vehicles (UAV), and Radio-frequency identification (RFID) sensors embedded in devices and cities have enabled an unprecedented increase in the amount of location records of moving objects on earth, such as taxi GPS traces and tourist check-ins. Such a series of temporally-ordered location points of an object represents a trajectory. Mining trajectory data is important to a broad range of applications such as location-based social networks, intelligent transportation systems, and urban computing \cite{zheng2015trajectory}. Trajectory data mining involves two important tasks: 1) Trajectory representation learning, which aims at encoding trajectory data  into  (low-dimensional)  vector  space; and  2)  Trajectory  generation, which  reversely  aims  at constructing a trajectory-structured data from low-dimensional space containing the trajectory generation rules or distribution. Different  from  trajectory  representation learning which benefits the downstream tasks such as discriminative learning and clustering, trajectory generation focuses on learning and interpreting the underlying distribution and mechanism of the trajectory generative process, which is crucial for tasks such as human mobility simulation and privacy preservation of individual traces~\cite{wang2020survey}. In this paper, we focus on the topic of trajectory data generation.
\begin{figure}
    \centering
    \includegraphics[width=\linewidth, trim=7.9cm 6.5cm 3cm 7cm, clip]{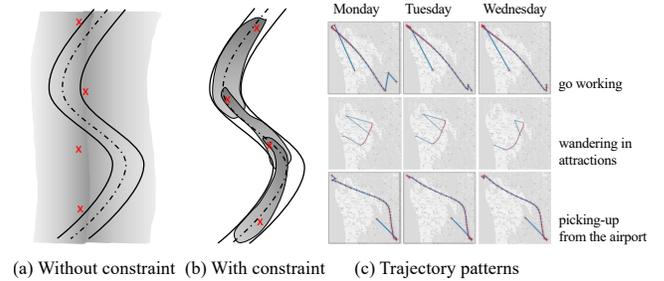}
    \caption{{\small (a) red crosses are points of a trajectory, while gray zone are the probability density envelop that a deep learning model without constraint has; (b) the probability density envelop (grey zone) with roads as constraints; (c) for different time of a day, there could be multiple choices of routing for the same start and end point in a city. Start and end is determined by users which is static patterns, but routing choices are dynamic patterns.}}
    \vspace{-16pt}
    \label{fig:examples}
\end{figure}

Traditional models for trajectory generation typically rely on are hand-crafted rules or prescribed distributions \cite{giannotti2005synthetic,pelekis2013hermoupolis} which require extensive human labors and domain knowledge yet still suffer from the bias and limited knowledge of the sophisticated mechanisms in trajectory generation. To address these issues, a fast-developing research topic is to extend deep generative models toward trajectory data, which enables to learn expressively generative models that could learn the sophisticated distributions based on a large amount of historical data. This is inspired by the success of deep generative neural networks in images and texts.
Although deep learning has been widely used for other trajectory data mining tasks, such as trajectory representation learning and prediction, deep generative models for trajectory generation have not been well explored as indicated in a recent survey \cite{wang2020survey}. This is a fast-growing domain, where existing few relevant works for trajectory generation are based on image-based Generative Adversarial Network (GAN) models \cite{ouyang2018non,smolyak2020coupled} and sequential Variational Autoencoder (VAE) \cite{huang2019variational}.

Despite the progress in this promising domain in recent years, there are still several important challenges yet to be addressed:
\textbf{Challenge 1: the necessity and difficulty in factorizing semantic and spatiotemporal patterns in trajectory generative modeling.} Each trajectory typically comes with an underlying purpose, such as ``go working'', ``wandering in attractions'', or ``picking-up people from the airport''. This type of patterns is namely global semantic meaning that do not change across different location points inside the trajectory. In addition, trajectory naturally also comes with local patterns that characterize the information for each location inside it as well as their spatiotemporal dependencies. Explicitly differentiating them has not been well explored by the existing deep generative models, which has limited the model interpretability and generalizability. \textbf{Challenge 2: Difficulty in jointly ensuring spatiotemporal-validity of the generated trajectories.} A generated trajectory is reasonable only when it satisfies necessary geometrical, physical, social principles. For example, all the location points should be on the roads and the movement speed should be limited to a reasonable range. Although deep generative models are good at learning expressive distributions from data, the learned distributions are still smoothed distribution over observations. Therefore, it is difficult yet imperative to diminish the probabilistic density for the invalid patterns. \textbf{Challenge 3: More reasonable inductive bias upon the prior distributions is needed.} Existing deep trajectory generative models usually follow the conventional priors used in deep generative models, which is to assume the independency among the latent variables corresponding to different locations. This, however, may not be ideal for trajectory generation because of the inherent dependence between the consecutive location points. How to design a new prior that goes beyond the conventional priors (e.g., isotropic Gaussians) is preferable yet challenging for trajectory generation. 

To address the above issues, we proposes a new framework of factorized deep generative models for trajectory generation with spatiotemporal-validity constraints. Through factorized latent variables, it speparates global semantics as well as local spatiotemporal semantic. Newly-generalized dependent priors for latent sequential variables are proposed contrast to conventional independent priors in sequential models. With a novel constrained optimization solution, it reduces the probability of generating invalid samples. Extensive experiments with ablation study and qualitative study showed the effectiveness of different latent variables and this constrained optimization.

\section{Related Work}
\textbf{Trajectory Generation/Synthesis:} 
This domain has a long history, where the representative methods include Oporto \cite{giannotti2005synthetic} based physical movement estimation, or Hermoupolis \cite{pelekis2013hermoupolis} based on urban points of interests. See \cite{wang2020survey} for a comprehensive survey. Such conventional methods are hard to replicate since it uses many ground features of a specific city, and requires extensive programming efforts and domain knowledge to implement. The current emerging trend for trajectory generation is to use deep generative models in a data-driven end-to-end fashion. Deep generative models for trajectory generation are not widely explored until now \cite{wang2020survey}. One type of works converted trajectories to images first and applied GANs for generation tasks \cite{ouyang2018non,smolyak2020coupled}. Such an approach loses many aspects of information including time, speed, and directions. Another work \cite{huang2019variational} utilize vanilla variational autoencoder scenarios by generating a whole trajectory via variables from unit Gaussian, which cannot jointly encode time-variant and -invariant information  \cite{kingma2013auto}. Deep generative models that can comprehensively take care of static and dynamic patterns in trajectory while ensuring the spatiotemporal validity are seriously under-explored yet imperative.

\textbf{Spatially-valid constraints in trajectory:} other studies on trajectories consider spatial-temporal-validity constraints, such as trajectory generation of vehicles \cite{choe2015trajectory}, collisions avoidance \cite{mehdi2017piecewise}, monitoring with turning constraints \cite{stephens2019randomized}, Trajectory tracking with velocity and heading rate constraints \cite{ren2004trajectory}, bounded zoning constraints \cite{jorris2007common}. Such constraints are not trivial to be considered in deep generative models and raise a major obstacle to generate realistic trajectory by neural networks. 

\textbf{Disentangled and factorized deep generative models:} Disentangled deep generative models are promising research topic recently, especially for applications on image data~\cite{bang2019explaining,chen2018isolating,higgins2016beta,kim2018disentangling}. The notion of disentanglement and factorization is to separate out the underlying explanatory factors responsible for variations of data. The generative representation learned in this way have been relatively resilient to the complex variants involved~\cite{bengio2013representation}, and can be used to enhance interpretability, generalizability, and robustness against adversarial attack~\cite{bang2019explaining}. Additional inductive bias could be considered to further factorize by leveraging particular data properties, such as factorizing graph data into the node and edge patterns~\cite{guo2020interpretable} and factorizing video data into the object and motion patterns~\cite{li2018disentangled}.

\section{Spatiotemporal-valid Trajectory Generation}
\label{sec:method}
We first introduce the Bayesian network of the proposed generative model, followed by new model inference methods. Then spatiotemporal-validity constraints are described and induced to the training objective. Finally, the model architectures for the encoder and decoder are elaborated.
\begin{figure}[!t]
    \centering
    \includegraphics[width=0.6\linewidth]{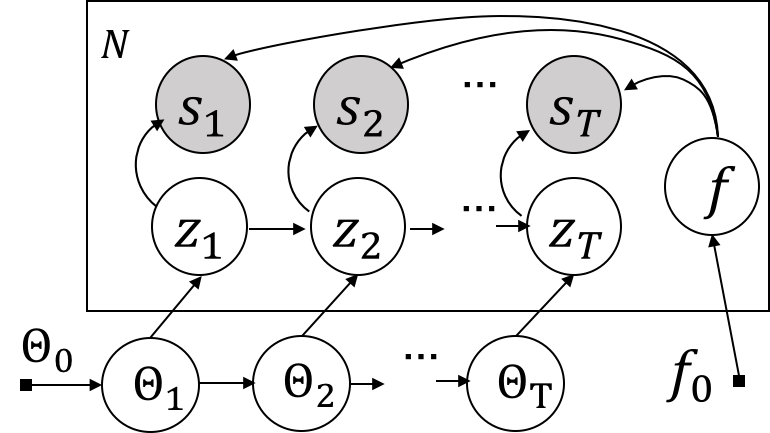}
    \caption{{\small Plate notation of the proposed deep generative models for generating $N$ trajectories. The index of the variables for each sample has been omitted for simplicity.}}
    \vspace{-12pt}
    \label{fig:graphical_models}
\end{figure}

\subsection{Generative model}
\label{sec:graphical_generative_model}
First, we define a trajectory as a sequence of location points $\{s_1,s_2,\cdots,s_T\}$ at time points $1,2,\cdots,T$. The proposed method focuses on a new generative process of trajectories, which factorizes the whole semantic meaning of a trajectory into two parts: 1) the global semantics of the whole trajectory as well as 2) local semantics that characterize the dependencies among the neighborhood. The global semantics cover the overall meaning of the trajectory including commuting from suburb to downtown, wandering inside downtown, jogging in the trails, and so on. The local semantics cover spatiotemporal autoregressive patterns such as how the next location is dependent on the current locations. Moreover, instead of assuming that the local semantic variable $\{z_1,\cdots,z_T\}$ must be all from identical and independent prior distributions, here we allow their priors $\{\Theta_1,\cdots,\Theta_T\}$ to be conditional dependent with each other. When $z_t$ follows Gaussian distribution, $\Theta_t=\{\mu_t,\sigma_t\}$ which are mean and standard deviation. These conditional distributions introduce more reasonable inductive bias and model expressiveness, by considering the dependencies among spatial neighbors. Specifically, as also shown in Figure \ref{fig:graphical_models}, our model characterizes the following generative process:

\begin{itemize}
\item Draw a sequence of the priors $\{\Theta_1,\cdots,\Theta_T\}$ for local semantic status, based on conditional probability: $\Theta_t\sim p(\Theta_t|\Theta_{t-1})$, where $p(\Theta_0)$ is a predefined distribution such as a unit Gaussian.

\item For each trajectory, draw a time-invariant variable $f$ as the global semantic from $p_{\theta}(f)$ such as a unit Gaussian;
\item For each trajectory, draw the local semantic variable $z_1$ for the first time point $t=1$ from $\Theta_1$.
\begin{itemize}
    \item For each time point $t\ge 1$, draw the underlying local semantic variable $z_t$ with the conditional probability $z_t\sim p(z_t|z_{t-1},\Theta_t)$.
    \item For each $t$, draw the observed variable $s_t\sim p(s_t|z_t,f)$.
\end{itemize}
\end{itemize}

A natural question is what the connection of our work and previous works in trajectory generation is. It is found that if $\mu_{1:T}, \sigma_{1:T}$ are generated from $f$ instead of $\pmb{0}$, our models collapse to the baseline SVAE model in \cite{huang2019variational} and $z_t,\mu_t, \sigma_t$ would become internal parameters and states that have no significant meanings. We also provide such ablation study in later experiment sections \ref{sec:experiment} to support the usage of dynamic factors $z_{1:T}$ with its priors $\mu_{1:T}, \sigma_{1:T}$.

\subsection{Model Inference}
\label{sec:training_objective}
Since the proposed generative model is intractable to infer, we proposed to solve it based on variational inference used for training variational autoencoder. This is achieved by first establish an approximate posterior $q_{\phi}(z_{1:T},f|s_{1:T})$ in order to approximate the original distribution $p(z_{1:T},f|s_{1:T})$, we investigate two possible choices of $q_{\phi}$:
\begin{equation}\begin{aligned}\nonumber
q_{\phi}(f, z_{1:T}| s_{1:T}) = 
\begin{cases}
q_{\phi}(f | s_{1:T}) q_{\phi}(z_{1:T},\Theta_{1:T} | s_{1:T})\ \  (\textit{factorized})\\
q_{\phi}(f | s_{1:T}) q_{\phi}(z_{1:T},\Theta_{1:T} |f, s_{1:T})\ \ \ \ \ \ \  (\textit{full})
\end{cases}
\end{aligned}\end{equation}
where the level of variance of $z_{1:T}$ could change depending on $f$ in full model, for example, if most roads between home and work are highways, then there is almost no variance for routing choice, while the level of noise of $z_{1:T}$ in the factorized model do not depend on $f$. Such modeling could reflect on different road network layout of different cities. 

Following $\beta$-VAE, the objective is as follows:
\begin{align}
\min_{\psi,\phi} \mathcal{L}(p_{\psi}, q_{\phi})=-&\mathbb{E}_{q_{\phi}}\left[\log p_{\psi}(s_{1:T}|z_{1:T},f,\Theta_{1:T})\right]\\\nonumber
+ \beta {KL}(q_{\phi}&(z_{1:T},f,\Theta_{1:T}|s_{1:T})||p_{\psi}(z_{1:T},f,\Theta_{1:T}))
\end{align}
where $\beta$ is hyper-parameter to control disentanglement in $\beta$-VAE, $KL$ is shorten for Kullback–Leibler divergence \cite{higgins2016beta}, $\psi$ and $\phi$ are sets of parameters in neural networks. They could be the parameters of a predefined distribution or deep generative neural networks. The first term is typically used for minimizing the reconstruction loss while the second one helps regularize the learned posterior close to the prior distributions. More specifically, the second term can be expanded as follows:
\begin{align}
\label{eq:obj_fun}
    &{KL}(q_{\phi}(z_{1:T},f,\Theta_{1:T}|s_{1:T})||p_{\psi}(z_{1:T},f,\Theta_{1:T}))\\\nonumber
    = & {KL}\left(q_{\phi}(z_{1:T},f,\Theta_{1:T}|s_{1:T})||p(f)\prod_{t=1}^Tp(z_{t}|z_{t-1},\Theta_t)p(\Theta_t|\Theta_{<t})\right)
\end{align}
where the prior $p(\Theta_0)$ follows an unit Gaussian distribution.

\subsection{Spatiotemporal-validity constraints}
\label{sec:spatial_constraint_reformulation}
Although the generative model learned by the Equation \ref{eq:obj_fun} could effectively characterize the underlying process of trajectory generation, the trajectories sampled from the learned generative model may not guarantee its validity and physical meaning in the real world. For example, the probabilistic density of the trajectory usually is continuous in the whole geographical space, leaving any location with non-zero probability to be passed by the trajectory. However, a trajectory needs to strictly follow spatial constraints. For example, the trajectory of vehicles needs to be on the roads, and hence its shapes and patterns should be constraints by the geometry of the roads. This requires to diminish the probabilistic density for the unfeasible trajectory patterns such as ``out of road'' or ``constantly back and forth''. Embedding in such an inductive bias can effectively increase the model generalizability and possibly strengthen the robustness against noise in training data due to the inaccuracy of the sensing data (e.g., those from GPS). The notion of spatial validity constraints can be leveraged our Equation \ref{eq:obj_fun}, by the newly extended objective:

The central contribution is imposing spatial validity constrains in optimizing generic VAE loss function $\mathcal{L}$ that we have developed in Equation \ref{eq:obj_fun} as follows:
\begin{align}
\label{eq:contrained_objective}
\min\nolimits_{\psi, \phi} \mathcal{L}(p_{\psi}, q_{\phi}), \,\, s.t. \forall s_{1:T}\notin\mathcal{C}: p_{\psi}(s_{1:T}|z_{1:T},f,\Theta_{1:T})=0
\end{align}
where $\mathcal{C}$ denotes the set of all the trajectory patterns that satisfy the spatial validity constraint. The spatial validity constraint can be specified by the user based on the practical need. For example, if the constraint says all the locations in the trajectory must be on the roads, then $\mathcal{C}_1=\{[x_1,\cdots,x_T]|x_t\in \mathcal R\}$, where $\mathcal R$ denotes the spatial regions of the roads. The constraint could also be on the first-order phenomena such as speed limit, meaning the trajectory's moving speed must be physically feasible for the moving object. This could be denoted as  $\mathcal{C}_2=\{[x_1,\cdots,x_T]||\Delta x_t|\le\mathcal S\}$, where $|\Delta x_t|$ denotes the object's speed at time $t$ while $\mathcal S$ is the speed limit that this object's speed cannot exceed. Another pattern could be the turning angles between two consecutive segments in the trajectory, in many situations, it is unlikely to have many consecutive sharp turnings. To constrain this, we could have $\mathcal{C}_3=\{[x_1,\cdots,x_T]|\sum\nolimits_t cos(x_{t-1}-x_{t},x_{t+1}-x_t)<\lambda\}$, where $cos(x_{t-1}-x_{t},x_{t+1}-x_t)$ denotes the cosine similarity of the two vectors each of which is the movement in the 2D Euclidean surface. The spatial constraint $\mathcal C$ can also be composed of the logical combinations among multiple rules. For example, $\mathcal C=\mathcal C_2 \bigcap(\mathcal C_1  \bigcup\mathcal C_3)$.

Directly solving complex constrained problems using conventional ways such as Lagrangian has been demonstrated to be inefficient for deep neural networks. Here we extend a recent deep constrained optimization framework \cite{ma2018constrained} to handle our problem in Equation \ref{eq:contrained_objective}, which is reformulated as follows:
\begin{align}\nonumber
\label{eq:regularize_objective}
\tilde{\mathcal{L}}(p_{\psi}, q_{\phi},\gamma)&= \mathcal{L}(p_{\psi}, q_{\phi})+\gamma\left[\right.\int \mathbbm{1}(g(z_{1:T},f,\Theta_{1:T})\notin\mathcal{C}) \\
&\cdot p_{\psi}(z_{1:T},f,\Theta_{1:T})\ \mathrm{d}z_{1:T}\ \mathrm{d}f\ \mathrm{d}\Theta_{1:T}\left.\right]^{\frac{1}{2}}
\end{align}
where $\mathcal{C}$ is the set of validity functions, and $\mathbbm{1}(\cdot)$ is an indicator function that output $1$ if a generated trajectory is invalid, otherwise $0$. We can reduce the integral term with a common approach of Monte Carlo Sampling in VAE \cite{kingma2013auto}. To allow gradient-flow over the regularization term, constraint functions in $\mathcal{C}$ must have gradients.

\subsection{Deep neural network architectures}
\label{sec:neural_network_architectures}
In this section, we introduce the detailed architectures for our proposed STG. Let a trajectory $s_{1:T}$ in our database $\mathcal{S}$, and $s_t = \langle x_1^t, x_2^t \rangle$ denotes the $t$th coordinate at time step $t$. The abstracted operations are shown in Figure \ref{fig:framework} with sub-modules. Our encoder is $q_{\phi}(z_{1:T},f,\Theta_{1:T}|s_{1:T})=q(z_{1:T}\Theta_{1:T}|f, s_{1:T})q(f|s_{1:T})$, which can be decomposed into two sub-encoders. 1) \textit{time-invariant encoder $q(f|s_{1:T})$}, which consumes the whole sequence that capture the stochastic whole-sequence representation $f$ detailed in upper left corner of Figure \ref{fig:framework}; 2) \textit{time-variant encoder}, with factorized modeling alternative $q(z_{1:T}\Theta_{1:T}|s_{1:T})$ and full modeling alternative $q(z_{1:T}\Theta_{1:T}|f, s_{1:T})$, which takes each coordinates step by step to generates a stochastic posterior representation $z_t$ for each step detailed in lower left corner of Figure \ref{fig:framework}. Blue lines are for full modeling alternative; 3) \textit{joint-factor decoder for training $p_{\psi}(s_{1:T}|z_{1:T},f,\Theta_{1:T})$} during training phrase that combine sampled $y$ and $z_t$ to stochastically generate each coordinates $s_t$ step by step, and minimize our training loss, which is detailed in right part of Figure \ref{fig:framework}. For \textit{joint-factor decoder for synthesis}, joint-factor decoder relies only on the sequential network to generate prior means and variances of $z_t$ first without the need to use encoder. In the following, we use $MLP_*(\cdot)$ to denote different multi-layer perceptron, use $BiLSTM_*$ for different bi-directional LSTM \cite{graves2005framewise}, use $RNN_*$ for vanilla RNN networks. In general, all the operations are as follows:
\begin{figure}[!t]
    \centering
    \includegraphics[width=\linewidth, trim=13.5cm 7.5cm 0cm 3cm, clip]{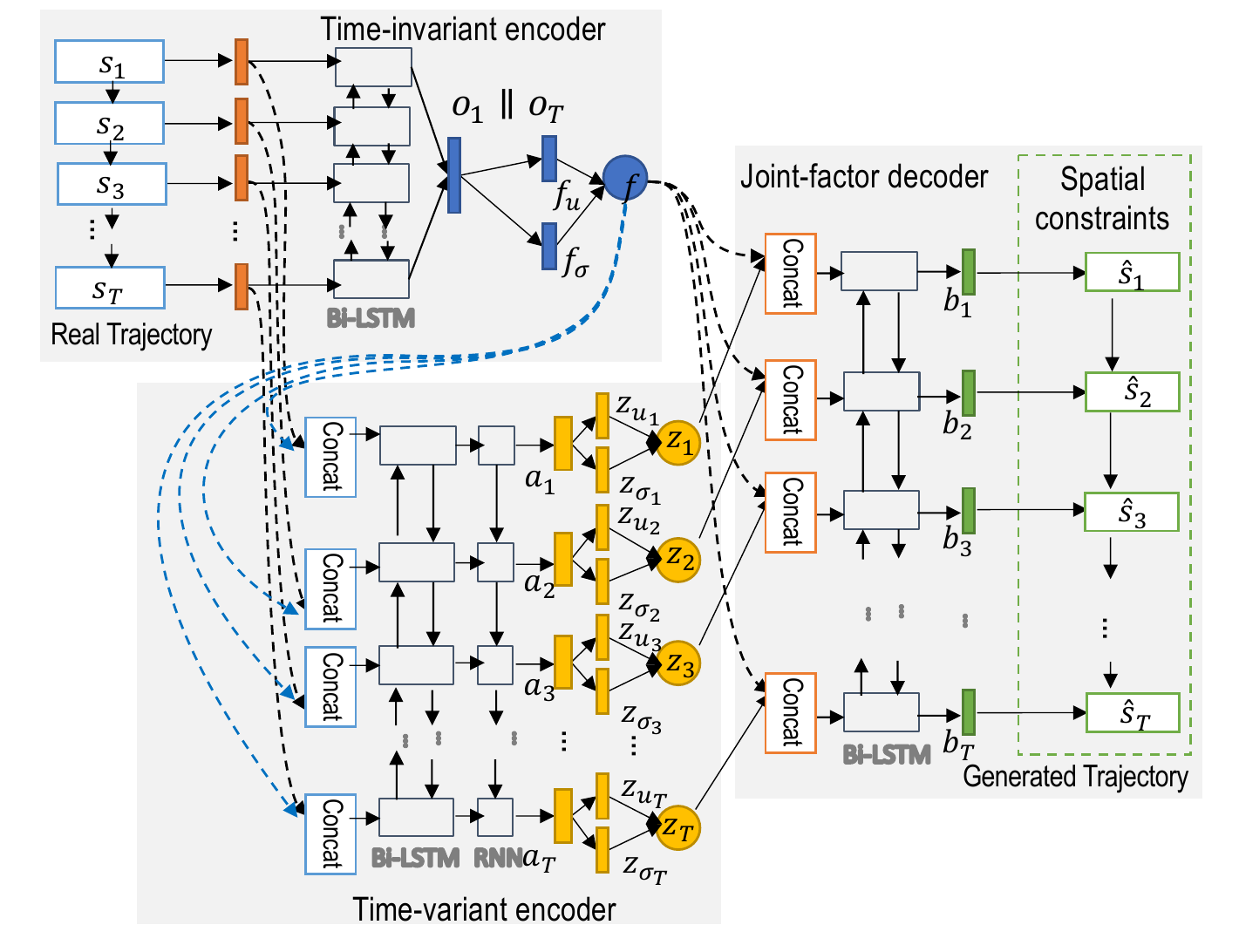}
    \caption{{\small spatiotemporal-valid Trajectory Generative Architect.}}
    \vspace{-12pt}
    \label{fig:framework}
\end{figure}
\begin{figure}[!t]
    \centering
    \includegraphics[width=\linewidth, trim=13.5cm 8cm 0cm 4cm, clip]{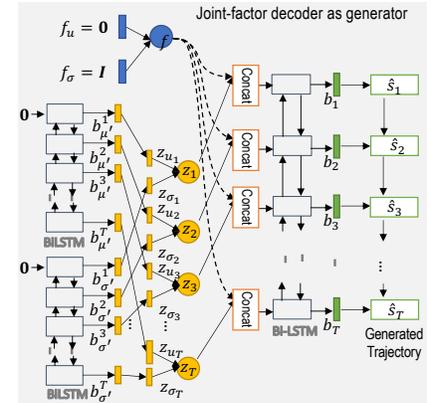}
    \caption{{\small Joint-factor decoder for synthesis: a dynamic sequential generator to sample from sequential meta-priors, which make $z_{1:T}$ not i.i.d. samples but with recurrent dependence.}}
    \vspace{-12pt}
    \label{fig:synthesis_generator}
\end{figure}

\textit{\textbf{Time-invariant encoder:}}
\small
$$m^{t+1}_f; o^{t+1} = BiLSTM_f(MLP_s(s_t), m^t_f, o^t)$$
$$\mu_f = MLP_{\mu_f}(o^1 || o^T), \,\sigma_f = MLP_{\sigma_f}(o^1 || o^T),\, y \sim \mathcal{N}(\mu_f, \sigma_f)$$\normalsize

\textit{\textbf{Time-variant encoders:}}
\small
$$\textit{(1) factorized:}\,\,\,\, {\Tilde m^{t+1}_z}; {\Tilde a^{t+1}} = BiLSTM_z(MLP_s(s_t), {\Tilde m^{t}_z}, {\Tilde a^{t}})$$
$$\textit{(2) full:}\,\,\,\, {\Tilde m^{t+1}_z}; {\Tilde a^{t+1}} = BiLSTM_z(MLP_s(s_t)||y, {\Tilde m^{t}_z}, {\Tilde a^{t}})$$
$$m^{t+1}_z; a^{t+1} = RNN_z({\Tilde m^{t+1}_z}, m^{t}_z; a^{t})$$
$$\mu_{z_t} = MLP_{\mu_{z_t}}(a^{t}), \sigma_{z_t} = MLP_{\sigma_{z_t}}(a^{t}),\,\,\,\, z_t \sim \mathcal{N}(\mu_{z_t}, \sigma_{z_t})$$\normalsize

\textit{\textbf{Joint-factor decoder for training:}}
\small
$$m^{t+1}_s; b^{t+1} = BiLSTM_s(y||z_t, m^t_s, b^t), \,\,{\widehat s_t} = MLP_s(b^t)$$\normalsize

\textit{\textbf{Joint-factor decoder for synthesis:}}
\small
$$m^{t+1}_{\mu}; b^{t+1}_{\mu} = BiLSTM_{\mu}(\pmb{0}, m^t_{\mu}, b^t_{\mu}), \,\, \mu_t = MLP_{\mu}(b^{t}_{\mu})$$
$$m^{t+1}_{\sigma}; b^{t+1}_{\sigma} = BiLSTM_{\sigma}(\pmb{0}, m^t_{\sigma}, b^t_{\sigma}),  \,\, \sigma_t = MLP_{\sigma}(b^{t}_{\sigma})$$
$$z_t \sim \mathcal{N}(\mu_t, \sigma_t)$$
$$m^{t+1}_s; b^{t+1} = BiLSTM_s(y||z_t, m^t_s, b^t), \,\,{\widehat s_t} = MLP_{b}(b^t)$$\normalsize
where $||$ is the concatenation operation of vectors, $\sim$ is the sampling operation which use the re-parameterization trick \cite{kingma2013auto} to allow gradient back-propagation. $m_*$ are different latent states vectors, and $o_*$, $a_*$, $b_*$ are outputs for either $BiLSTM_*$ or $RNN_*$ modules.\normalsize

\section{Experiments}
\label{sec:experiment}
In this section, both quantitative and qualitative results are reported to show the performance of STG with ablation study and comparisons to previous methods over four datasets. 
All experiments are conducted on a 64-bit machine with a NVIDIA 1080ti GPU.

\subsection{Datasets}
\subsubsection{Real-world datasets}
The first dataset is collected from $442$ taxi at Porto, Portugal describing a complete year (from 01/07/2013 to 30/06/2014) \footnote{{\scriptsize \url{ www.kaggle.com/c/pkdd-15-predict-taxi-service-trajectory-i/data}}}. Data do not have time-stamps but with a fixed $15$ second sampling interval. The second dataset is T-Drive data that collect continuous GPS points of 10,357 taxis in one week with real timestamps \footnote{{\scriptsize \url{www.microsoft.com/en-us/research/publication/t-drive-trajectory-data-sample/}}}. Preprocessing steps are used to clean the data, including Noise Filtering and Stay Point Detection \cite{zheng2015trajectory}. The third real-world dataset is human check-ins collected from a location-based website Gowalla \footnote{{\scriptsize \url{snap.stanford.edu/data/loc-Gowalla.html}}}, for which only the dense region at Dallas metropolitan from original global data is selected. All points are projected to a local geographic coordinate system in meters and convert to a 1000-meter unit.

\subsubsection{Synthetic dataset}
We generated a synthetic dataset for 10,000 students living on a university campus from a location-based simulator \cite{kim2020location}. The student agents mimic real-world contacting and check-ins patterns based on predefined living and social preference settings in agent-based simulation.

We convert all datasets to Euclidean space using geographically projection based on the original earth projection system used in data. We split raw data with 0.9/0.1 ratios for training and testing subsets.

\subsection{Constraints settings}
\subsubsection{Physics-induced constraint: }
Many constraints can be developed from physics law and engineering of a car. Since sampling time intervale $e$ is small (15 seconds), a constraint is that if the average speed $\gamma_t = \frac{||s_t - s_{t-1}||_2}{e}$ is higher than a threshold ${\bar \gamma} = 60 Km/h$, it is impossible for a car to make an sharp turn, so we can not observe a preceding angle $\eta_t$ (in cosine value) smaller than a threshold ${\bar \eta}$. And, $\eta_t = \frac{(s_t - s_{t-1}) \cdot (s_{t-1} - s_{t-2})}{||s_t - s_{t-1}||_2 ||s_{t-1} - s_{t-2}||_2}$. This regularization only impose penalties over a case that the angle is smaller than threshold $\eta_t > {\bar \eta}$ when a segment is larger than threshold $\gamma_t > {\bar \gamma}$ at the same time. We show such patterns in Porto dataset (red dashed region in second row) in Figure \ref{fig:constraint_compare} which is formulated as follows:
\small
\begin{equation}\begin{aligned}\nonumber
\label{eq:physics_constraint}
 \frac{\lambda}{N} \sum\nolimits^J \sum\nolimits_{t=2}^T c(t, s_{1:T}) &= \frac{\lambda}{N} \sum\nolimits^J  \sum\nolimits_{t=2}^T (\gamma_t - {\bar \gamma})_+ (\eta_t - {\bar \eta})_+
\end{aligned}\end{equation}\normalsize
where $s_t$ is treated as a vector, and $\lambda$ is a hyper-parameter because there is only one constraint function. Notice that total $T-2$ constraints for each trajectory are possible.\normalsize

\subsubsection{Behavior-induced constraints: } 
\begin{figure*}[!t]
    \centering
    \includegraphics[width=\linewidth, trim=1cm 9cm 1cm 6cm, clip]{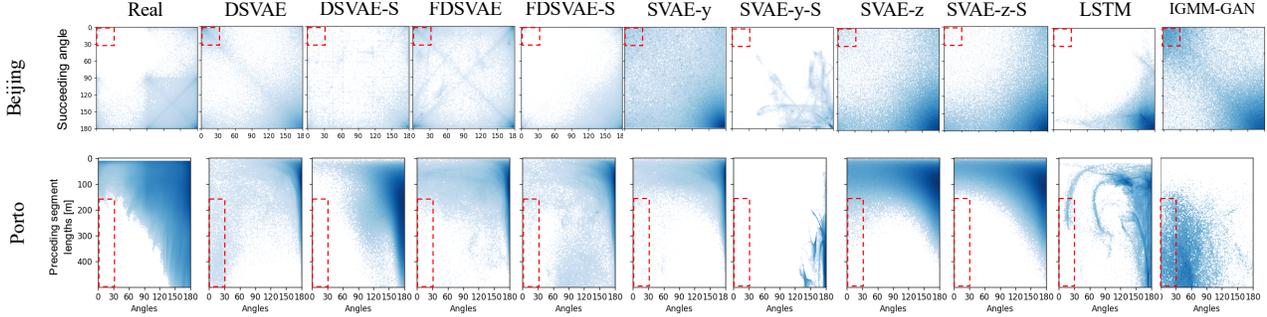}
    \caption{Spatial constraint results of Beijing and Porto datasets with red-dashed area for violation zone.}
    \vspace{-12pt}
    \label{fig:constraint_compare}
\end{figure*}

Behavioral constraints come from behaviors which are abnormal to human, even though these behaviors did not validate the physics laws. For example, in two consecutive segments with a high GPS sampling rate like $5$ second, it is abnormal to have two U-Turns (turn to opposite direction), in other words, two consecutive angles could not both be very sharp (less than 30 degrees). Such constraint can be shown in the Beijing dataset (red dashed regions in first row) in Figure \ref{fig:constraint_compare}. This regularization penalizes the case that first angle is smaller than a sharp threshold $\eta_t > {\bar \eta}$ when its preceding angle is also very sharp $\eta_{t-1} > {\bar \eta}$. Its formula is as follows:
\small\begin{equation}\begin{aligned}\nonumber
\label{eq:behavior_constraint}
\frac{\lambda}{N} \sum\nolimits^J \sum\nolimits_{t=3}^T c(s_{1:T})  &= \frac{\lambda}{N} \sum\nolimits^J \sum\nolimits_{t=3}^T (\eta_t - {\bar \eta})_+ (\eta_t - {\bar \eta})_+
\end{aligned}\end{equation}
where $\eta$ is also the cosine value of angles. Notice that total $T-3$ constraints for each trajectory are possible.\normalsize
\subsection{Competing methods and Ablation study}
Here, we introduce competing methods. Since some competing methods also belong to the ablation study, we also introduce the ablation study models simultaneously.

LSTM: a basic LSTM model that can take any start point as input and output a sequence of points.

IGMM-GAN: a GAN-based model with a new Dirichlet Process Mixture Model for latent noise input. It processes a trajectory as an image not a sequence \cite{smolyak2020coupled}.

SVAE-y: it used a static latent variable for a whole sequence, firstly developed in \cite{huang2019variational}. This method can be also treated as an ablation study for our model that we use $y$ without $z_{1:T}$.

SVAE-z: this ablation model uses only $z_{1:T}$ without $y$.

Disentangled SVAE (DSVAE): this is for the full model with both $y$ and $z_{1:T}$ and each $z_t$ is dependent on $y$.

Factorized Disentangled SVAE (FDSVAE): this is for the factorized alternative that each $z_t$ is independent on $y$.

Also, to provide ablation study to our other contribution, for each of the variantional based model (namely SVAE-y, SVAE-z, DSVAE, FDSVAE), we implement another version with spatial constraints namely SVAE-y-S, SVAE-z-S, DSVAE-S, FDSVAE-S. For human check-in trajectories, we do not do experiments with spatial constraints since such constraints do not hold for our datasets. Any other potential constraints are left to future works.

\subsection{Evaluation metrics:}
Mean Distance Error (MDE) in Haversine Distance or Euclidean distance is the most used metric in current works \cite{ouyang2018non,huang2019variational,smolyak2020coupled}, defined as $\frac{1}{n} \sum^N_i ||s_i - {\hat s_i}||_2$, where $||\cdot||_2$ denotes L2 norm. Since MDE only compute a reconstruction loss, we proposed to directly evaluate spatial feature distributions in ``Maximum Mean Discrepancy (MMD)'', a sophisticated distance function used in recent approaches in image \cite{zhang2020maximum}, text \cite{guo2020multi}, and graph \cite{you2018graphrnn}. A trajectory could have a feature vector $d_{1:T-1}$, and an original feature set is represented as $D \in \mathbf{R}^{N \times T}$, and $\Tilde{D}$ for a generated feature set. Function $MMD(D, \Tilde{D}) \mapsto [0, 1]$ has output value between $1$ for least similarity and $0$ for two exactly same distributions. The chosen spatial features include angles, segment lengths, total lengths, point counts of cells in a grid.

Violation Score (VS) is used to evaluate spatial constraint results. It's a ratio of violation case number over the total number of all cases. The lower the VS value is, the better a model outputs spatial-temporal-valid results. The formula is as follow:
\small
$$VS = \frac{\sum^N_i\sum^T_{t=t_*} \pmb{1}_{c}(t, s_{1:T}^{(i)})}{N \times (T - t_*)},
\pmb{1}_{c}(\cdot) := 
\begin{cases}
1 & \text{if    } c_*(\cdot) \wedge \dots \\
0 & \text{else}
\end{cases}
$$\normalsize
where $t_*$ is the start step defined for $c_*(\cdot)$, for example, $t_* = 2$ for segment length, while $t_* = 3$ for the angle of consecutive segments.\normalsize

\subsection{Evaluation results}
\label{sec:quantitative_results}
\textbf{General performances:} The performances of previous methods, our proposed STG methods, and ablated methods are presented in Table \ref{tab:mmd_result}. 
It gives the MDE score between a real trajectory and its reconstructed trajectory and compares angle distribution, segment length distribution, total length distribution, and grid point count distribution in MMD between real trajectory sets and synthetic sets. 
\begin{figure*}[!t]
    \centering
    \includegraphics[width=\linewidth, trim=0cm 9cm 0cm 6cm, clip]{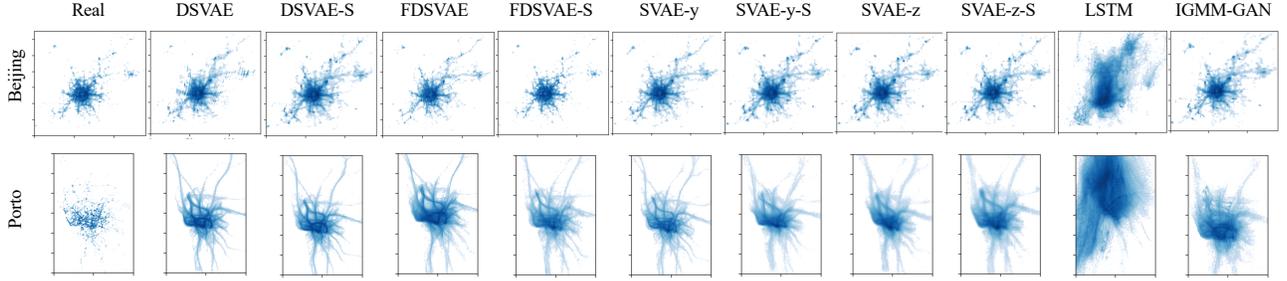}
    \caption{{\small Point count distributions over grids for synthetic trajectories and real taxi GPS trajectories comparisons in cities.}}
    \vspace{-12pt}
    \label{fig:reconstruct_whole_city}
\end{figure*}
\begin{figure}[!t]
    \centering
    \includegraphics[width=\linewidth, trim=0.5cm 9.5cm 6cm 6.5cm, clip]{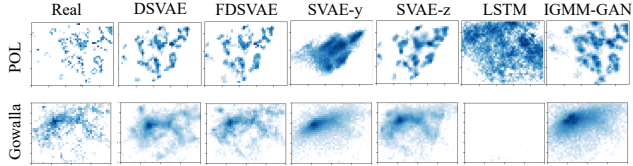}
    \caption{{\small Point count distributions over grids for synthetic trajectories and real check-ins comparisons.}}
    \vspace{-12pt}
    \label{fig:reconstruct_whole_city_checkin}
\end{figure}
\begin{table}[!t]
\scriptsize
    \centering
    \begin{tabular}{p{0.15in} p{.55in} p{0.2in} p{0.3in} p{.4in} p{.4in} p{.4in}}
    \toprule
        dataset & \diagbox[width=.6in]{Method}{Metrics}  & MDE & Angles in MMD & Segment lengths in MMD & Total lengths in MMD & Grid point counts in MMD \\
        \midrule
        \multirow{5}{*}{Porto} 
        & LSTM & 13.6525 & 0.5243 & 0.4976 & 0.4136 & 0.1135 \\
        & IGMM-GAN & 11.1488 & 0.0772 & 1.0 & 0.3779 & 0.0429 \\
        & SVAE-y & 1.2422 & 0.0430 & 0.0034 & 0.0655 & 0.0244 \\
        \cline{2-7}
        & SVAE-z & 1.73782 & 0.0081 & 0.0069 & 0.3303 & 0.0279 \\
        & DSVAE & 0.9018 & 0.0649 & 0.0041 & 0.2439 & 0.0208 \\
        & FDSVAE & 1.7415 & 0.0380 & 0.0019 & 0.0497 & 0.0294 \\
        & SVAE-y-S & 1.9755 & 0.0795 & 0.1009 & 0.6947 & 0.0647 \\
        & SVAE-z-S & 1.1896 & \textbf{0.0054} & 0.0055 & 0.2593 & 0.0106 \\
        & DSVAE-S & \textbf{0.8059} & 0.0628 & 0.0032 & 0.2133 & 0.0208 \\
        & FDSVAE-S & 1.2281 & 0.0273 & \textbf{0.0004} & \textbf{0.0495} & \textbf{0.0105} \\
        \hline
        \multirow{5}{*}{Beijing} 
        & LSTM & 16.0594 & 0.4005 & 0.6447 & 0.4080 & 0.3717 \\
        & IGMM-GAN & 0.9577 &  0.0556 & 0.7280 & 0.1052 & 0.003 \\
        & SVAE-y & 0.6073 & 0.3844 & 0.3563 & 0.1409 & 0.1360 \\
        \cline{2-7}
        & SVAE-z & 0.9849 & 0.0178 & 0.0074 & 0.0437 & 0.0005 \\
        & DSVAE & 0.5916 & 0.1310 & 0.0788 & 0.1448 & 0.1040 \\
        & FDSVAE & 1.1136 & 0.3635 & 0.0076 & 0.7308 & 0.1594 \\
        & SVAE-y-S & 1.5179 & 0.4316 & 0.3456 & 0.1288 & 0.1207 \\
        & SVAE-z-S & 0.9138 & 0.0200 & 0.0079 & \textbf{0.0436} & \textbf{0.0002} \\
        & DSVAE-S & \textbf{0.5130} & \textbf{0.0047} & 0.0087 & 0.0852 & 0.0088 \\
        & FDSVAE-S & 0.6118 & 0.0088 & \textbf{0.0018} & 0.0610 & 0.0660 \\
        \hline
        \multirow{5}{*}{POL} 
        & LSTM & 34.5681 & 0.5921 & 0.9292 & 0.9915 & 0.9835 \\
        & IGMM-GAN & \textbf{0.8872} & 0.1957 & \textbf{0.0003} & 0.0004 & \textbf{0.0023} \\
        & SVAE-y & 10.6391 & 0.2107 & 0.6874 & 1.0067 & 0.4426 \\
        \cline{2-7}
        & SVAE-z & 6.1959 & 0.1629 & 0.0044 & 0.0012 & 0.0480 \\
        & DSVAE & 8.5842 & 0.1012 & 0.0030 & \textbf{0.0003} & 0.0463 \\
        & FDSVAE & 7.2282 & \textbf{0.0935} & 0.0094 & 0.0009 & 0.0456 \\
        \hline
        \multirow{5}{*}{Gowalla} 
        & LSTM & 629.07 & 0.0325 & NaN & NaN & NaN \\
        & IGMM-GAN & 101.32 & 0.0351 & 0.1015 & 0.0431 & 0.0163 \\
        & SVAE-y & 90.292 & 0.0111 & 0.0430 & 0.0017 & 0.0030 \\
        \cline{2-7}
        & SVAE-z & 40.2241 & \textbf{0.0067} & 0.0328 & 0.0002 & 0.0026 \\
        & DSVAE & 2.7645 & 0.0065 & 0.0196 & 0.0002 & 0.0036 \\
        & FDSVAE & \textbf{1.7714} & 0.0102 & \textbf{0.0144} & \textbf{0.0001} & \textbf{0.0025} \\
        \bottomrule
    \end{tabular}
    \caption{{\small Experiment results.}}
    \label{tab:mmd_result}
\end{table}

For the two taxi trajectory datasets, our STG outperformed other competing and ablated methods in most metrics. The margin of improvement of VAE-based models and IGMM-GAN compared to LSTM is huge. It is caused by the lack of randomness in vanilla LSTM. 
VAE-based models are overall preferred in MMDs. Specifically, comparing simple SVAE-y and SVAE-z to our proposed DSVAE and FDSVAE in Table \ref{tab:mmd_result}, DSVAE is the best in MDE for both taxi dataset. The spatial constraint versions normally improve over non-constraint ones, so DSVAE-S and FDSVAE-S gained the best performances in most metrics except SVAE-z-S method achieve slightly better in angles for Porto and in total lengths and grid points in MMD for Beijing. 
The overall grid point distributions in Figure \ref{fig:reconstruct_whole_city} also shows that all VAE-like models with/without constraints are generally better than LSTM and IGMM-GAN. 

For two check-in datasets in Table \ref{tab:mmd_result}, FDSVAE is preferred, except that IGMM-GAN is the lowest in MDE, segment length, and grid point of POL dataset. This might result from the relatively small sample size of POL. For the Gowalla dataset, FDSVAE won in almost all metrics, except SVAE-z won in angles. By showing grid point density in Figure \ref{fig:reconstruct_whole_city_checkin}, it is also confirmed that DSVAE, FDSVAE, and SVAE-z captured a similar pattern to real datasets.

\textbf{Constraint performances:} By comparing models without constraints with models with constraints in Table \ref{tab:constraint_result}, proposed spatial regularization terms help to generate much fewer violation cases for all models since VS consistently decrease after adding constraints. In the plotted distribution of related features in Figure \ref{fig:constraint_compare}, there is much more white space (indicating zero number of samples) in red dashed regions after adding constraints.

\begin{table}[!t]
\scriptsize
    \centering
    \begin{tabular}{p{0.2in}p{0.6in}p{.4in}p{0.6in}p{.4in}}
    \toprule
        dataset & Method & VS & Method & VS \\
        \midrule
        \multirow{5}{*}{Porto} 
        & plain LSTM & 0.045219 & - & - \\
        & IGMM-GAN & 0.02624 & - & - \\
        & SVAE-y & 0.034881 & SVAE-y-S & 0.004960\\
        & SVAE-z & 0.018214 & SVAE-z-S & 0.002749\\
        & DSVAE & 0.027971 & DSVAE-S & 0.003682 \\
        & FDSVAE & 0.021269 & FDSVAE-S & 0.001180 \\
        \cline{2-5}
        & Raw data & 0.001718 & - & - \\
        \hline
        \multirow{5}{*}{Beijing} 
        & plain LSTM & 0.004753 & - & - \\
        & IGMM-GAN & 0.055332 & - & - \\
        & SVAE-y & 0.008197 & SVAE-y-S & 0.033250\\
        & SVAE-z & 0.009581 & SVAE-z-S & 0.006895 \\
        & DSVAE & 0.010779 & DSVAE-S & 0.003263 \\
        & FDSVAE & 0.054197 & FDSVAE-S & 0.007847 \\
        \cline{2-5}
        & Raw data & 0.003395 & - & - \\
        \bottomrule
    \end{tabular}
    \caption{{\small Violation score Experiment results.}}
    \vspace{-12pt}
    \label{tab:constraint_result}
\end{table}


%
\subsection{Disentanglement analysis}
\label{sec:disentanglement_analysis}
\begin{figure}[!t]
    \centering
    \includegraphics[width=\linewidth, trim=0cm 0cm 0cm 0cm, clip]{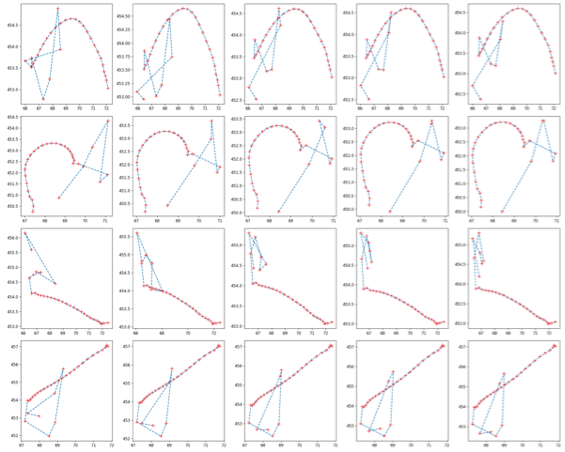}
    \caption{{\small Disentangled factors studies. x axis is to change time-variant $z_t$ factors, y axis is to change time-invariant $y$ factor. This experiment used Porto dataset with FDVAE-S model.}}
    \vspace{-16pt}
    \label{fig:disentangle_analysis}
\end{figure}

In this part, we demonstrate a qualitative analysis to show that the factorization of time-variant and time-invariant factors achieved better interpretability in Figure \ref{fig:disentangle_analysis}. We use FDVAE-S model with Porto dataset as an example. Along x-axis, the sampled $z_{1:T}$ vectors' second dimension is replaced with values from 1 to 9. Along y-axis, we randomly sampled 9 different $y$ vectors from distribution $\mathcal{N}(\pmb{0}, \pmb{1})$. We can see that $y$ controls the overall trend of trajectories since trajectories in the same row present highly similar patterns. The $z_t$ injected different noises to each step of trajectories, since trajectories in the same row show small variances. For the same column, it shows that $z_t$ might control some high-dimensional geometric dynamics, though it is hard to visually conclude any specific geometric factor that $z_t$ controls. 

\section{Conclusion}
\label{sec:conclusion}
We develop a novel STG framework for deep generative models with spatiotemporal-validity constraints, which achieved better performance not only in the conventional MDE metric but also over feature distribution in MMD metrics and violation score. It shows that the effectiveness of factorizing time-variant and time-invariant factors, sequential priors over each time step, and constrained optimization. Even though taxi GPS trajectories and check-ins trajectory are selected, our STG framework can be also applied for other similar data mining tasks, and in other domains like animal migration trajectory, ant movement trajectory, and sport trajectory, which are left to the future works.


\bibliography{main.bib}

\appendix

\subsection{Neural network details}
We introduce details of neural networks, especially number of neurons. For Porto and Beijing dataset, the $MLP_s(\cdot)$ is multiply layers with $[48, 16]$. For all $BiLSTM_*$ and $RNN_*$ modules, the dimensions of the hidden states (not shown in our paper) are set to $512$. For the $BiLSTM_f$ of static pattern, the dimension of recurrent input $o^t$ is $16$. The $MLP_{\mu_f}$ and $MLP_{\sigma_f}$ with inputs of $512 \times 2$ dimension, and with output dimension of $256$. For the $BiLSTM_z$ of dynamic patterns, the dimension of input ${\Tilde a}^t$ is $16$. The second $RNN_z$ module takes forward and backward hidden states as an input, whose dimension is $512 \times 2$. The hidden state of $RNN$ is with $512$ dimension.  $MLP_{\mu_{z_t}}$ and $MLP_{\sigma_{z_t}}$ is set to have one layer of $64$ neurons. The priors $\Theta$ includes $\mu_t$ and $\sigma_t$, whose decoding modules $BiLSTM_{\mu}$ and $BiLSTM_{\sigma}$ have the same design of $BiLSTM_z$. The difference is that its input is a $\pmb{0}$ vector with dimension of $16$. The $MLP_{\mu}$ and $MLP_{\sigma}$ have the same number of $64$ neurons. The $f||z_t$ input's dimension is $256 + 64$ for the decoder module $BiLSTM_s$. The $MLP_s$ has one internal layer of $128$ neurons and a last layer of two neurons for two coordinate values in $s_t$.

There are a few differences for POL and Gowalla data. The $MLP_s(\cdot)$ is multiply layers with $[48, 32]$. For all $BiLSTM_*$ and $RNN_*$ modules, the dimensions of the hidden states (not shown in our paper) are set to $512$. For the $BiLSTM_f$ of static pattern, the dimension of recurrent input $o^t$ is $32$. The $MLP_{\mu_f}$ and $MLP_{\sigma_f}$ with inputs of $512 \times 2$ dimension, and with output dimension of $256$. For the $BiLSTM_z$ of dynamic patterns, the dimension of input ${\Tilde a}^t$ is $32$. The second $RNN_z$ module takes forward and backward hidden states as an input, whose dimension is $512 \times 2$. The hidden state of $RNN$ is with $512$ dimension.  $MLP_{\mu_{z_t}}$ and $MLP_{\sigma_{z_t}}$ is set to have one layer of $32$ neurons. The priors $\Theta$ includes $\mu_t$ and $\sigma_t$, whose decoding modules $BiLSTM_{\mu}$ and $BiLSTM_{\sigma}$ have the same design of $BiLSTM_z$. The difference is that its input is a $\pmb{0}$ vector with dimension of $32$. The $MLP_{\mu}$ and $MLP_{\sigma}$ have the same number of $32$ neurons. The $f||z_t$ input's dimension is $256 + 32$ for the decoder module $BiLSTM_s$. The $MLP_s$ has two internal layer of $[64, 32]$ neurons and a last layer of two neurons for two coordinate values in $s_t$.

\subsection{Other parameter tuning}
Except changing different neural networks architectures, there are several hyper-parameters to be tuned, including: 1) $\beta$ parameter for $\beta$-VAE to enhance disentangling. We test values in $[1, 10, 100]$. $100$ is chose for the model in our paper; 2) $\gamma$ parameter for regularization of constraints. We test values in $[1, 10, 100]$. $1$ is chose for the presented model in the paper; 3) other conventional parameters. Learning rate is set to be $0.0002$ for Porto, $0.0002$ for Beijing, $0.0002$ for POL, and $0.002$ for Gowalla. The training epoch are all set to be $100$. The batch size is set to $128$ for all datasets and models. 

\subsection{Additional experiment results}
In this part, we did extensive case studies for different datasets so as to illustrate the effectiveness of our factorization approaches. Each row is generated with a fixed $f$, and each column is generated with a fixed $z_{1:T}$ sequence. We can see that for both taxi trajectories and check-in trajectories $f$ controls a static pattern (similar patterns in each row), while $z_t$ control the variances for each trajectory in such a row.

\begin{figure*}
    \centering
    \includegraphics[width=\linewidth]{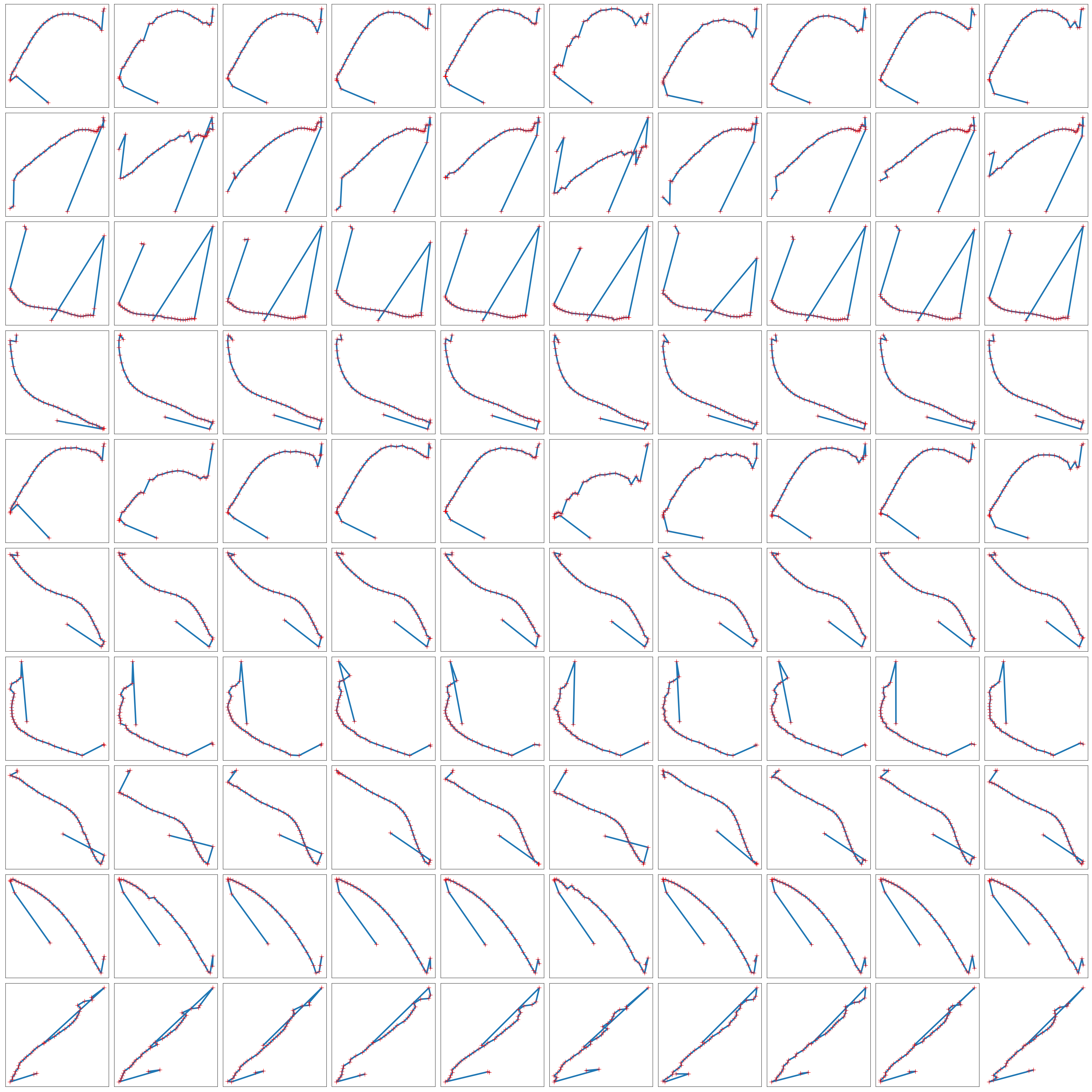}
    \caption{Case studies of $f$ and $z_t$ over Porto dataset}
    \label{fig:case_study_porto}
\end{figure*}

\begin{figure*}
    \centering
    \includegraphics[width=\linewidth]{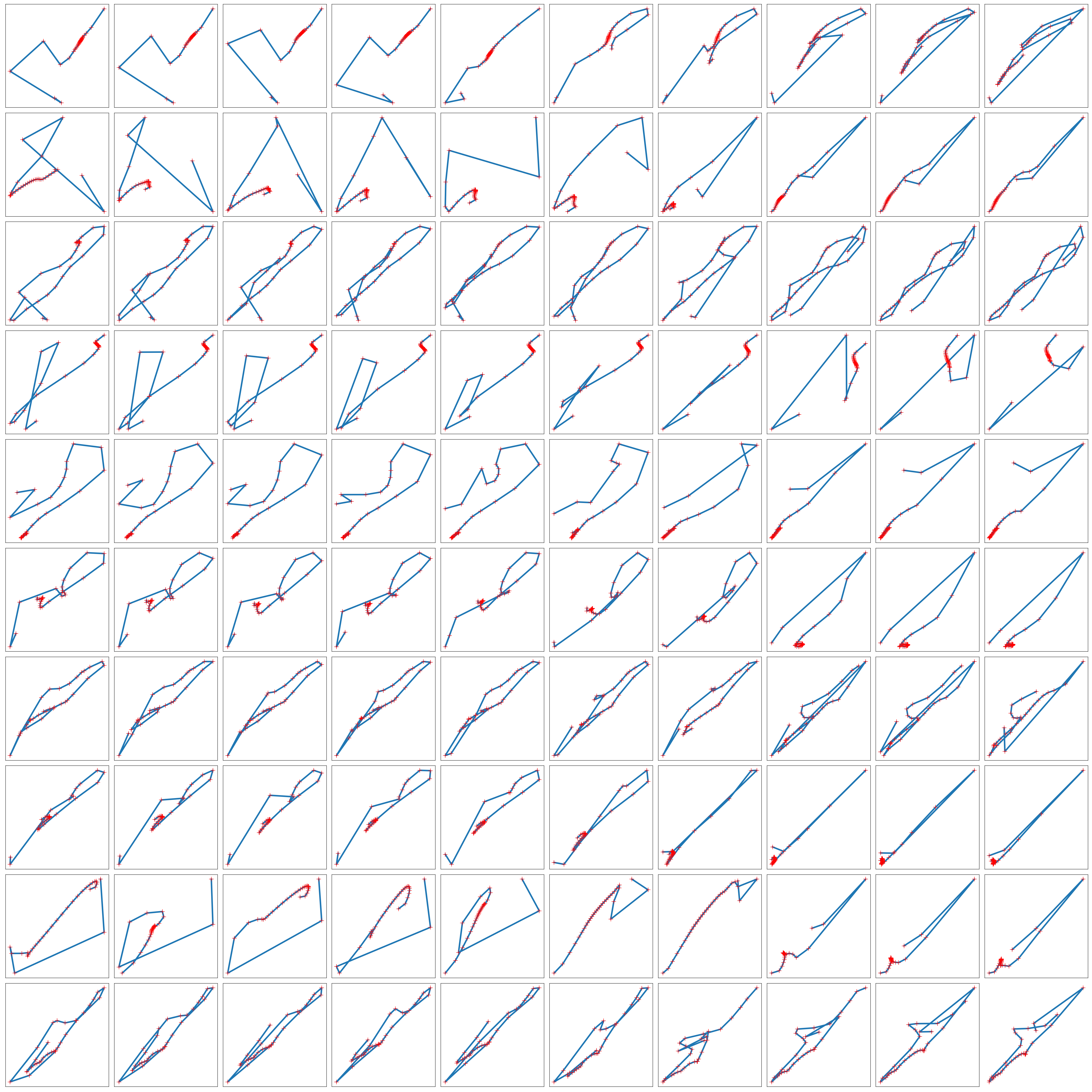}
    \caption{Case studies of $f$ and $z_t$ over Beijing dataset}
    \label{fig:case_study_beijing}
\end{figure*}

\begin{figure*}
    \centering
    \includegraphics[width=\linewidth]{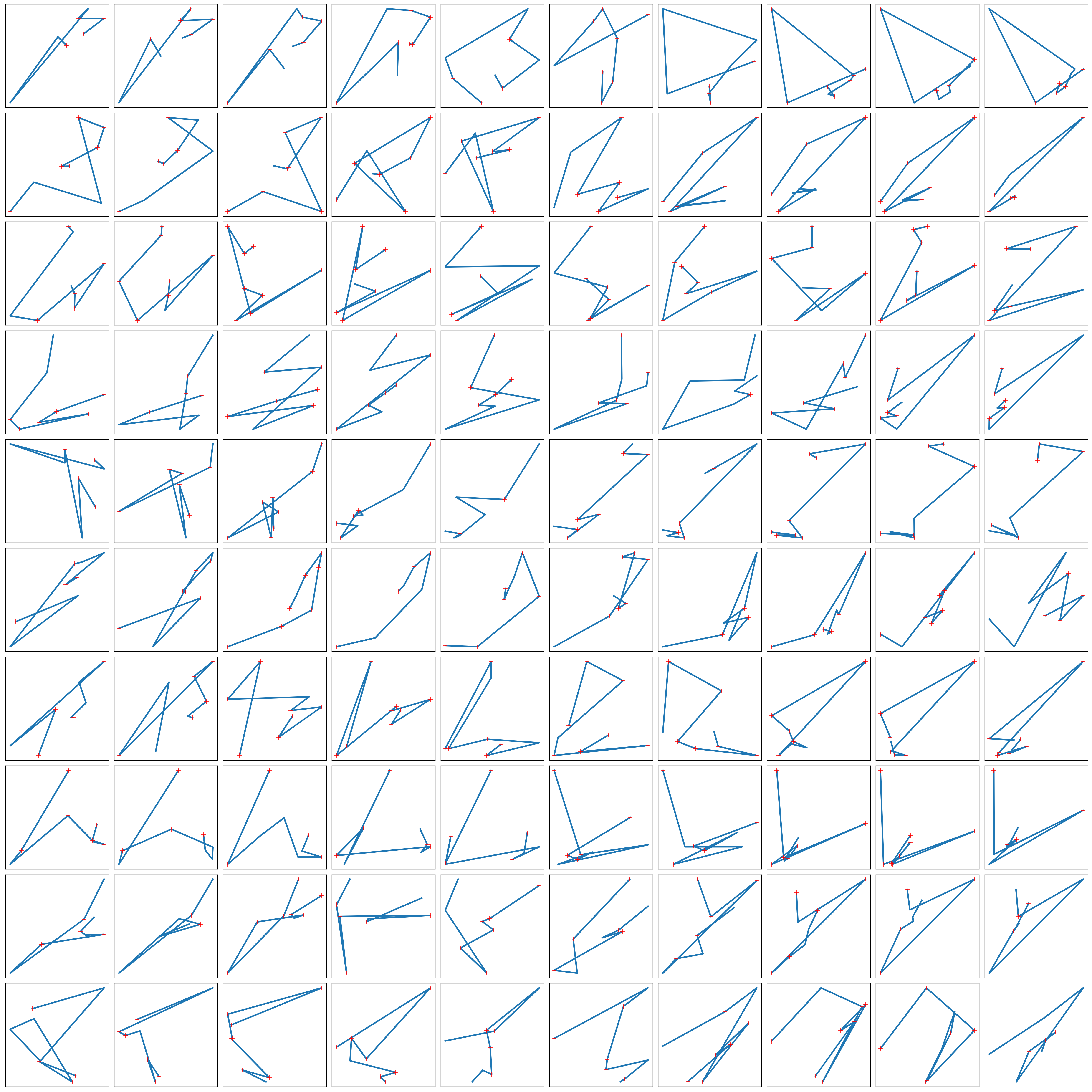}
    \caption{Case studies of $f$ and $z_t$ over POL dataset}
    \label{fig:case_study_pol}
\end{figure*}

\begin{figure*}
    \centering
    \includegraphics[width=\linewidth]{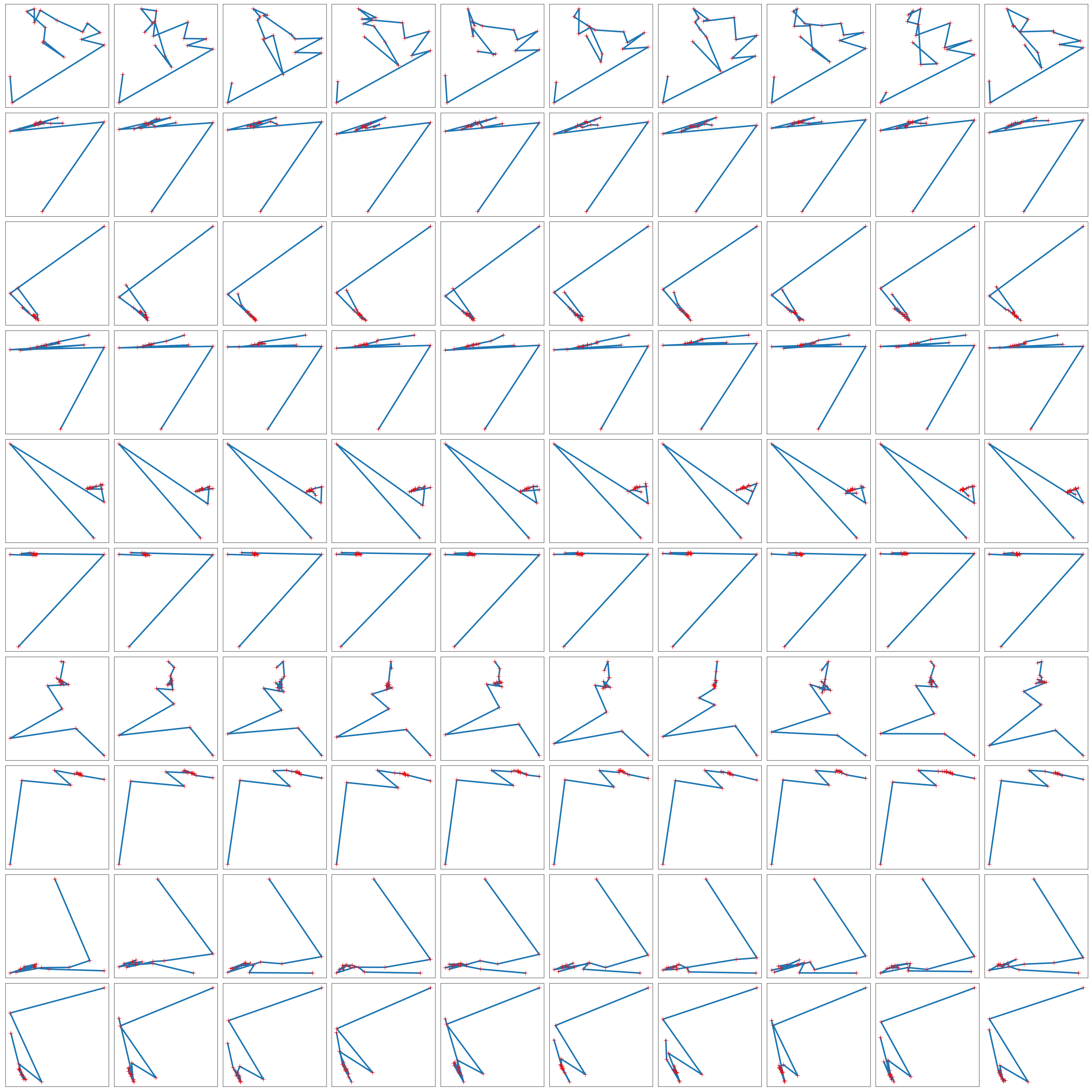}
    \caption{Case studies of $f$ and $z_t$ over Gowalla dataset}
    \label{fig:case_study_gowalla}
\end{figure*}

\end{document}